\documentclass[pdflatex,sn-mathphys-num]{sn-jnl}

\usepackage{graphicx}%
\usepackage{multirow}%
\usepackage{amsmath,amssymb,amsfonts}%
\usepackage{amsthm}%
\usepackage{mathrsfs}%
\usepackage[title]{appendix}%
\usepackage{xcolor}%
\usepackage{textcomp}%
\usepackage{manyfoot}%
\usepackage{booktabs}%
\usepackage{algorithm}%
\usepackage{algorithmicx}%
\usepackage{algpseudocode}%
\usepackage{listings}%
\usepackage{subcaption}
\usepackage{booktabs}
\usepackage{multirow}

\theoremstyle{thmstyleone}%
\theoremstyle{thmstyletwo}%

\theoremstyle{thmstylethree}%

\begin{document}

\title[Neural Network Assignment]{Performance Analysis of Image Classification on Bangladeshi Datasets}


\author*[1]{\fnm{Mohammed Sami} \sur{Khan}}\email{khanmsami@gmail.com}
\equalcont{These authors contributed equally to this work.}

\author*[1]{\fnm{Fabiha} \sur{Muniat}}\email{muniatfabiha999@gmail.com}
\equalcont{These authors contributed equally to this work.}

\author*[1]{\fnm{Rowzatul} \sur{Zannat}}\email{w.rzrowza@gmail.com}
\equalcont{These authors contributed equally to this work.}

\affil*[1]{\orgdiv{Department of Computer Science and Engineering}, \orgname{Dhaka University}, \orgaddress{ \city{Dhaka}, \postcode{1000},  \country{Bangladesh}}}


\abstract{Convolutional Neural Networks (CNNs) have demonstrated remarkable success in image classification tasks; however, the choice between designing a custom CNN from scratch and employing established pre-trained architectures remains an important practical consideration. In this work, we present a comparative analysis of a custom-designed CNN and several widely used deep learning architectures, including VGG-16, ResNet-50, and MobileNet, for an image classification task. The custom CNN is developed and trained from scratch, while the popular architectures are employed using transfer learning under identical experimental settings. All models are evaluated using standard performance metrics such as accuracy, precision, recall, and F1-score. Experimental results show that pre-trained CNN architectures consistently outperform the custom CNN in terms of classification accuracy and convergence speed, particularly when training data is limited. However, the custom CNN demonstrates competitive performance with significantly fewer parameters and reduced computational complexity. This study highlights the trade-offs between model complexity, performance, and computational efficiency, and provides practical insights into selecting appropriate CNN architectures for image classification problems.}




\maketitle

\section{Introduction}\label{sec1}

In recent years, deep learning has emerged as a dominant paradigm in computer vision, with Convolutional Neural Networks (CNNs) \cite{cnn} playing a central role in achieving state-of-the-art performance across a wide range of image classification tasks. By exploiting local spatial correlations through convolutional operations, CNNs are capable of learning hierarchical feature representations that progress from low-level visual patterns, such as edges and textures, to high-level semantic concepts. As a result, CNN-based models have been successfully deployed in numerous real-world applications, including medical image analysis, autonomous driving, remote sensing, and biometric recognition \cite{app}.

Traditionally, CNNs have been designed and trained from scratch for specific tasks, requiring careful architectural design and extensive hyperparameter tuning. While custom CNN architectures offer flexibility and task-specific optimization, they often demand large amounts of labeled data and substantial computational resources to achieve competitive performance. In practice, many real-world datasets are limited in size, making it challenging for custom-designed models to generalize effectively without overfitting. These challenges have motivated the growing adoption of transfer learning techniques, where models pre-trained on large-scale datasets are reused for downstream tasks.

Transfer learning has significantly influenced modern deep learning workflows by enabling the reuse of rich feature representations learned from massive datasets such as ImageNet. Popular CNN architectures, including VGG-16 \cite{vgg}, ResNet-50 \cite{restnet}, and MobileNet \cite{mobilenet}, have become widely used benchmarks due to their proven performance, architectural innovations, and availability of pre-trained weights. VGG-16 introduced a simple yet effective deep architecture based on stacked convolutional layers, ResNet-50 addressed the degradation problem in deep networks through residual connections, and MobileNet emphasized computational efficiency using depthwise separable convolutions. These models represent different design philosophies, balancing depth, accuracy, and computational cost.

Despite the widespread success of pre-trained CNN architectures \cite{pretrainedcnn}, the choice between employing a custom-designed CNN and adopting a pre-trained model remains an important consideration. Custom CNNs may offer advantages in terms of architectural simplicity, reduced parameter count, and lower computational overhead, making them suitable for resource-constrained environments. Conversely, pre-trained models often achieve superior performance due to their ability to leverage prior knowledge, particularly when labeled data is scarce. A systematic comparison between these approaches is therefore essential to understand their respective strengths, limitations, and practical trade-offs.

Several studies have explored the effectiveness of transfer learning \cite{vgg} in image classification tasks; however, many works focus primarily on pre-trained models without providing a detailed comparison against custom architectures designed specifically for the target dataset. Moreover, differences in experimental settings, data preprocessing techniques, and evaluation protocols often make it difficult to draw fair and reproducible conclusions. To address these limitations, a controlled and comprehensive evaluation under identical conditions is necessary.

In this paper, we present a comparative analysis of a custom Convolutional Neural Network designed and trained from scratch and several widely adopted pre-trained CNN architectures, namely VGG-16, ResNet-50, and MobileNet, for an image classification task. All models are trained and evaluated using the same dataset, preprocessing pipeline, and evaluation metrics to ensure a fair comparison. Performance is assessed using standard classification metrics, including accuracy, precision, recall, and F1-score, alongside considerations of computational complexity and training efficiency.

The main contributions of this work can be summarized as follows:
\begin{enumerate}[label=(\alph*)]
\item We design and implement a custom CNN architecture tailored for the given image classification task.

\item We apply transfer learning using popular pre-trained CNN architectures under identical experimental settings.

\item We conduct an extensive comparative evaluation to analyze performance, convergence behavior, and computational trade-offs.

\item We provide practical insights to guide model selection for image classification problems based on dataset size and resource constraints.
\end{enumerate}
The remainder of this paper is organized as follows. Section 2 reviews related work on custom CNN architectures and transfer learning-based approaches. Section 3 describes the dataset and preprocessing steps. Section 4 presents the proposed methodology, including the custom CNN design and pre-trained models. Section 5 details the experimental setup, while Section 6 discusses the results and analysis. Finally, Section 7 concludes the paper and outlines directions for future research.

\section{Methodology}\label{sec2}

\subsection{Dataset Description}\label{subsec1}
This part discusses the datasets used to evaluate the custom CNN model against two pretrained models.

\subsubsection{Auto-RickshawImageBD}
This dataset \cite{autorickshaw} contains 1331 annotated images captured in different urban traffic scenes in Bangladesh. The images are annotated to distinguish auto rickshaws from non-auto-rickshaws, which makes the dataset suitable for binary classification.

\subsubsection{FootpathVision}
FootpathVision \cite{footpathvision} is an image dataset for the detection of sidewalk encroachments in urban areas. It contains 1238 images of various types of encroachments such as parked vehicles, shops, and other obstructing objects in pedestrian pathways. The dataset supports binary classification task distinguishing encroached vs. unencroached footpaths to improve pedestrian safety.

\subsubsection{RoadDamageBD}
RoadDamgeBD \cite{road} dataset has 450 images labeled as damaged roads and good-conditioned roads. Its purpose is to enable binary classification that supports city management. Using the augmentation technique, this dataset may perform better in Convolutional Neural Network (CNN) models.

\subsubsection{MangoImageBD}
MangoImageBD \cite{mangoimagedb} is an image dataset consisting of 15 classes representing 15 different mango varieties found in Bangladesh. The full dataset has total 28515 images with a combination of real and virtual backgrounds. There are also augmented images which enhances the datasets ability for computer vision tasks.

\subsubsection{PaddyVarietyBD}
PaddyVarietyBD \cite{paddyvarietybd} is a standard and open dataset of paddy varieties grown in Bangladesh. The dataset contains 14,000 RGB microscopic images of different paddy kernels, labeled for multiclass classification to support variety identification and agricultural decision-making. 35 unique paddy varieties invented in Bangladesh have been selected for this study.

\subsection{Model Description}\label{subsec2}
This part contains descriptions about the custom CNN architecture and the pretrained models used.
\subsubsection{Custom Model}
The classification task in this study is performed using a custom-designed Convolutional Neural Network (CNN) optimized for image data resized to 512 × 512 pixels. The model is built using PyTorch and follows a hierarchical feature-extraction approach, where spatial information is progressively compressed while deeper layers learn increasingly abstract representations. The custom CNN model consists of four sets of convolutional layers, each consisting of a $3\times3$ kernel with a stride of 1 and a padding of 1. Each convolutional layer is followed by a batch normalization layer into a ReLU activation, which is then followed by a Max Pool layer. The details about the model architecture can be found in Table ~\ref{tab:model-summary}.

\begin{table}[h!]
\centering
\caption{Summary of the custom CNN architecture and output dimensions.}
\begin{tabular}{l c c}
\hline
\textbf{Layer} & \textbf{Output Size} & \textbf{Parameters} \\
\hline
Input Image & $3 \times 512 \times 512$ & -- \\

Conv2D (3, 32, 3×3) + ReLU & $32 \times 512 \times 512$ & 896 \\
MaxPool2D (2×2) & $32 \times 256 \times 256$ & 0 \\

Conv2D (32, 64, 3×3) + ReLU & $64 \times 256 \times 256$ & 18{,}496 \\
MaxPool2D (2×2) & $64 \times 128 \times 128$ & 0 \\

Conv2D (64, 128, 3×3) + ReLU & $128 \times 128 \times 128$ & 73{,}856 \\
MaxPool2D (2×2) & $128 \times 64 \times 64$ & 0 \\

Conv2D (128, 256, 3×3) + ReLU & $256 \times 64 \times 64$ & 295{,}168 \\
MaxPool2D (2×2) & $256 \times 32 \times 32$ & 0 \\

Flatten & $262{,}144$ & 0 \\

Linear (256×256×32×32 → 512) & 512 & 134{,}218{,}240 \\
ReLU & 512 & 0 \\
Dropout (0.5) & 512 & 0 \\

Linear (512 → 3 classes) & 3 & 1{,}539 \\
\hline
\textbf{Total Parameters} & & \textbf{134,607,191} \\
\hline
\end{tabular}

\label{tab:model-summary}
\end{table}

Due to the task being classification, the model includes a fully-connected layer before the classification head. The model was designed with the following considerations:
\begin{itemize}
    \item Four convolutional blocks provide sufficient representational power without exceeding training resource limits.
    \item Batch Normalization and ReLU activation improves training stability and accelerates convergence.
    \item MaxPooling Downsampling reduces computational cost while preserving important spatial features.
    \item Dropout in Classifier adds regularization to reduce overfitting, especially important for moderate-sized datasets.
    \item Fully Connected Classifier provides non-linear decision boundaries and learns global combinations of local features.
    
\end{itemize}

\subsubsection{Pre-trained Model}
Two pre-trained convolutional architectures, ResNet-50 and ConvNeXt-Tiny \cite{convexttiny}, were used as fixed feature extractors to utilize knowledge learned from large-scale visual datasets.

ResNet-50 is a convolutional neural network consisting of 50 layers, trained on ImageNet to address the problem of vanishing gradients using residual connections \cite{he2016deep}. The model turned as a standard pre-trained backbone by achieving improved gradient flow and top performance on ImageNet. On the other hand, ConvNeXt-Tiny is a smaller model pre-trained on ImageNet. The architecture uses Vision Transformers-inspired design concepts while maintaining the inductive biases of convolutional neural networks \cite{liu2022convnet}.

Both models were initialized with ImageNet pre-trained weights to use as a feature extractor in this study. To maintain the general visual representations acquired from large-scale natural images, the backbone network's layers were all frozen during training. The frozen backbone had been added with a task-specific classification head. Depending on whether the classification is binary or multiclass, the final output layer employs either a sigmoid or a softmax function.
\subsubsection{Transfer Learning}
Building on the pre-trained configurations, the higher layers of the ResNet-50 and ConvNeXt-Tiny backbones were selectively fine-tuned to execute a transfer learning technique. The upper layers of the backbone network were unfrozen, but the bottom convolutional layers remained frozen. This selective unfreezing allows the model to modify high-level semantic characteristics while maintaining robust low-level feature representations. Fine-tuning was done to ensure consistent optimization. This method intends to improve the classification performance by enabling the model to efficiently learn domain-specific features.

\subsection{Hyperparameter Settings}\label{subsec3}
The custom convolutional neural network (CNN) was trained using a set of empirically chosen hyperparameters designed to balance training stability, convergence speed, and generalization performance. The training was performed for a maximum of 30 epochs, with early stopping applied based on validation loss to prevent overfitting. The Adam optimizer was used with the default values for $\beta_1$ (0.9) and $\beta_2$ (0.999) with a learning rate of $1\times10^{-5}$. A Cross-Entropy Loss was used, which is appropriate for multi-class image classification tasks. A batch size of 32 was used which fit into the 15 GiB VRAM provided by Kaggle.

To improve generalization, random transformations were applied during training: Random horizontal flip, Random rotation (±10°), Random resized crop, Normalization using ImageNet mean and standard deviation. These augmentations address dataset variability and mitigate overfitting. A dropout layer with a probability of 0.5 was included before the final fully connected layer to reduce co-adaptation of neurons and improve generalization performance. Training was monitored using validation loss. If the validation loss did not improve for 5 consecutive epochs, training was halted and the model parameters from the epoch with the lowest validation loss were restored.

\section{Experimental Results}\label{sec4}
Our custom CNN model was trained on the five datasets with slight modifications to the hyperparameters. The changes were mainly in batch size, maximum number of epochs, and learning rate. The results of training and evaluating on the five reported datasets are described in this section.

\subsection{Auto-RickshawImageBD}
\begin{figure}[h]
    \centering
    \begin{subfigure}[b]{0.48\textwidth} 
        \centering
        \includegraphics[width=\textwidth]{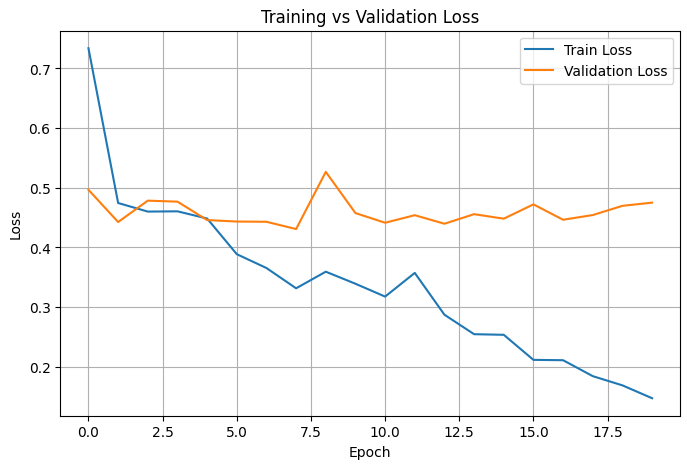}
        \caption{Custom CNN Loss} 
    \end{subfigure}
    \hfill
    \begin{subfigure}[b]{0.48\textwidth}
        \centering
        \includegraphics[width=\textwidth]{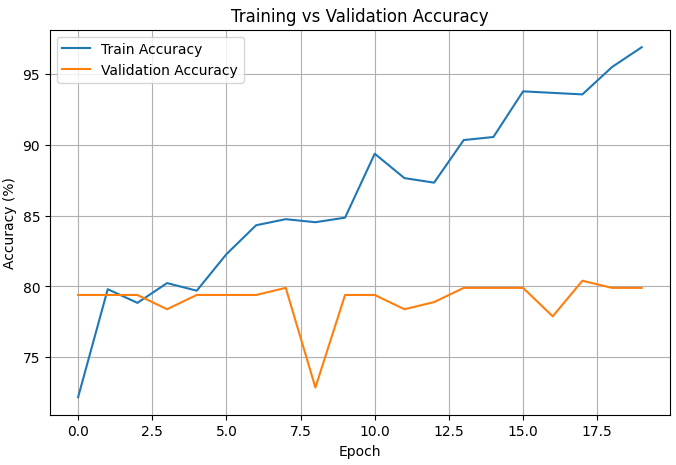}
        \caption{Custom CNN Accuracy}
    \end{subfigure}
    \caption{Training loss and Acuuracy curves of Custom CNN model across epochs on Auto-RickshawImageBD Dataset.}
    \label{fig:autorickshaw}
\end{figure}

\begin{figure}[h]
    \centering
    \begin{subfigure}[b]{0.46\textwidth} 
        \centering
        \includegraphics[width=\textwidth, height=4.5cm]{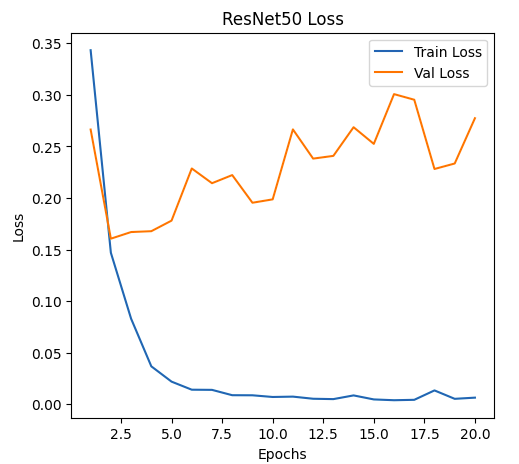}
        \caption{RestNet-50 Loss} 
    \end{subfigure}
    \hfill
    \begin{subfigure}[b]{0.46\textwidth}
        \centering
        \includegraphics[width=\textwidth, height=4.5cm]{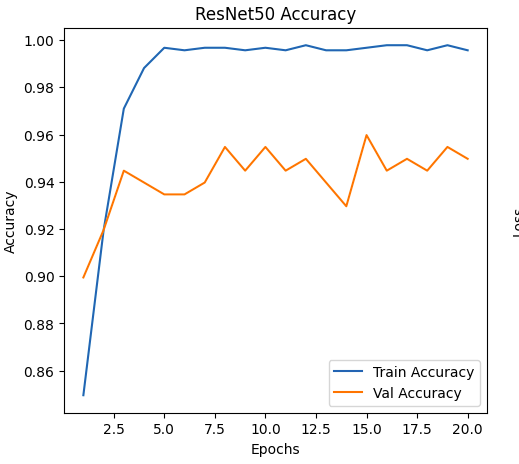}
        \caption{RestNet-50 Accuracy}
    \end{subfigure}
    \caption{Training loss and Acuuracy curves of RestNet-50 across epochs on Auto-RickshawImageBD Dataset.}
    \label{fig:autorickshaw_Restnet}
\end{figure}

\begin{figure}[h]
    \centering
    \begin{subfigure}[b]{0.46\textwidth} 
        \centering
        \includegraphics[width=\textwidth, height=4.5cm]{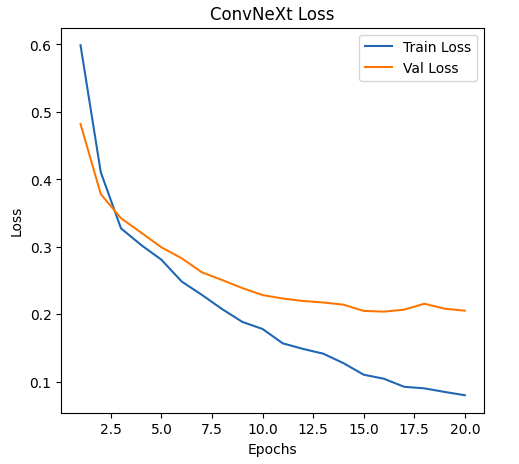}
        \caption{ConvNext-Tiny Loss} 
    \end{subfigure}
    \hfill
    \begin{subfigure}[b]{0.46\textwidth}
        \centering
        \includegraphics[width=\textwidth, height=4.5cm]{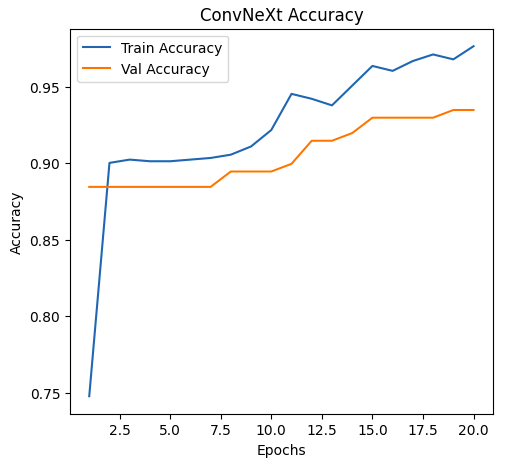}
        \caption{ConvNext-Tiny Accuracy}
    \end{subfigure}
    \caption{Training loss and Acuuracy curves of ConvNext-Tiny across epochs on Auto-RickshawImageBD Dataset.}
    \label{fig:autorickshaw_conv}
\end{figure}

The dataset \cite{autorickshaw} being designed for object detection, there is limited scope for classification. The data was loaded as a binary classification data with the positive class being the presence of an auto-rickshaw and the negative class being the absence thereof. With this settings, training was conducted and the data is given in Figure~\ref{fig:autorickshaw}, Figure~\ref{fig:autorickshaw_Restnet} and Figure~\ref{fig:autorickshaw_conv}.

For the Auto-RickshawImageBD dataset, the pretrained models converge substantially faster than the custom CNN, as indicated by the steeper decline in training loss and earlier stabilization of accuracy. This behavior reflects the effectiveness of transfer learning in capturing generic object-level features such as shape, edges, and vehicle structures. ConvNeXt-Tiny shows smoother convergence and stable accuracy trends, suggesting improved robustness to variations in viewpoint and background clutter. 
 

\subsubsection{FootpathVisionBD}
\begin{figure}[h]
    \centering
    \begin{subfigure}[b]{0.48\textwidth} 
        \centering
        \includegraphics[width=\textwidth]{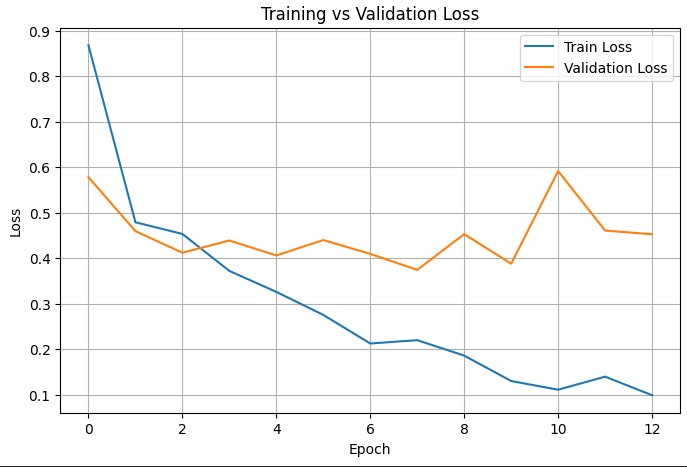}
        \caption{Custom CNN Loss} 
    \end{subfigure}
    \hfill
    \begin{subfigure}[b]{0.48\textwidth}
        \centering
        \includegraphics[width=\textwidth]{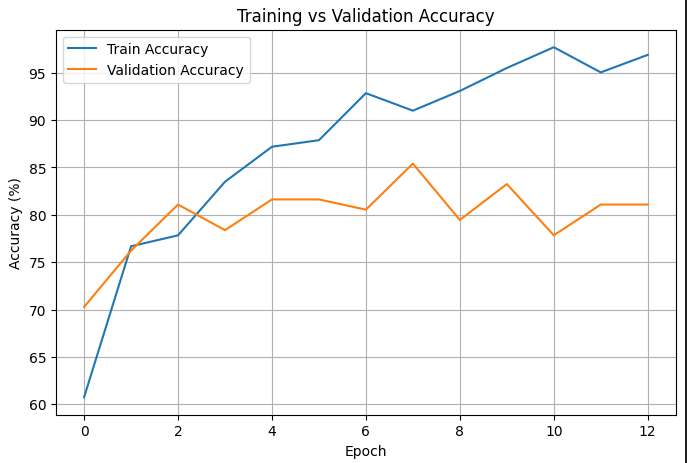}
        \caption{Custom CNN Accuracy}
    \end{subfigure}
    \caption{Training loss and Acuuracy curves of Custom CNN model across epochs on Auto-FootpathVisionBD Dataset.}
    \label{fig:foothpath}
\end{figure}

\begin{figure}[h]
    \centering
    \begin{subfigure}[b]{0.48\textwidth} 
        \centering
        \includegraphics[width=\textwidth, height=4.5cm]{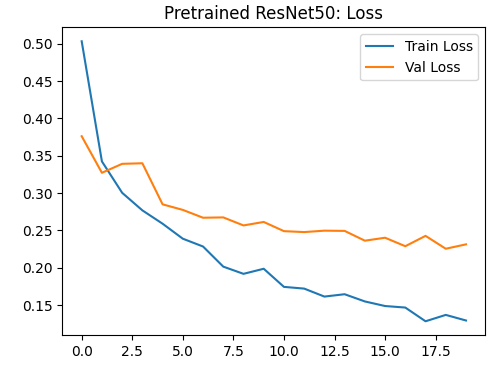}
        \caption{RestNet-50 Loss} 
    \end{subfigure}
    \hfill
    \begin{subfigure}[b]{0.48\textwidth}
        \centering
        \includegraphics[width=\textwidth, height=4.5cm]{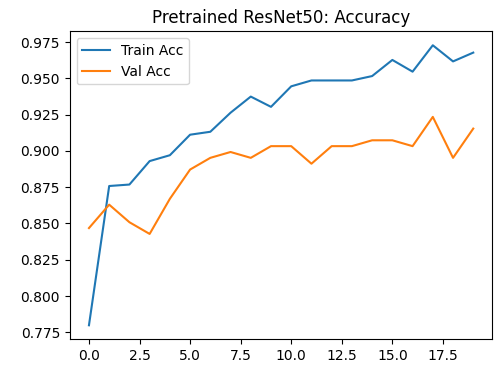}
        \caption{RestNet-50 Accuracy}
    \end{subfigure}
    \caption{Training loss and Acuuracy curves of RestNet-50 across epochs on Auto-FootpathVisionBD Dataset.}
    \label{fig:footpath_Restnet}
\end{figure}

\begin{figure}[h]
    \centering
    \begin{subfigure}[b]{0.48\textwidth} 
        \centering
        \includegraphics[width=\textwidth, height=4.5cm]{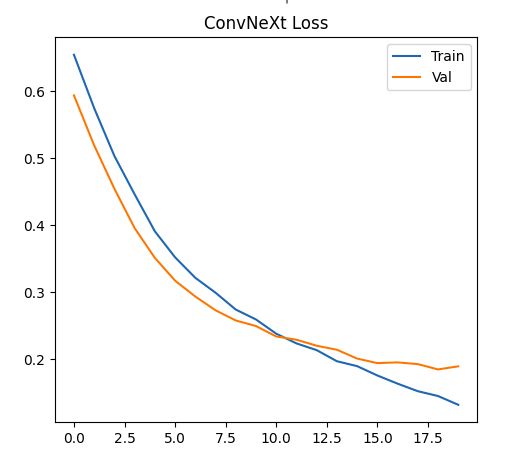}
        \caption{ConvNext-Tiny Loss} 
    \end{subfigure}
    \hfill
    \begin{subfigure}[b]{0.48\textwidth}
        \centering
        \includegraphics[width=\textwidth, height=4.5cm]{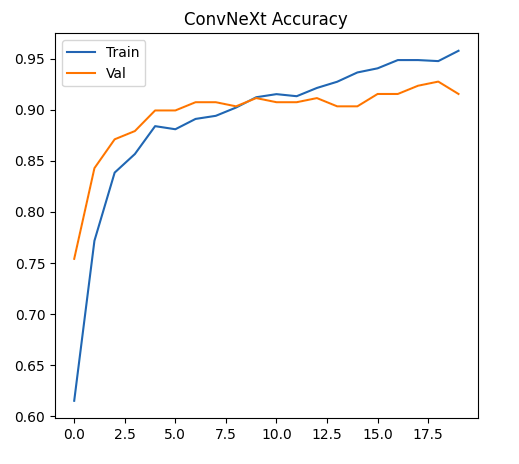}
        \caption{ConvNext-Tiny Accuracy}
    \end{subfigure}
    \caption{Training loss and Acuuracy curves of ConvNext-Tiny across epochs on Auto-FootpathVisionBD Dataset.}
    \label{fig:footpath_conv}
\end{figure}

The dataset was already designed for binary classification and after training for 20 epochs, 85.03\% accuracy was reached. The loss curve and accuracy curves are given in Figure \ref{fig:foothpath}, \ref{fig:footpath_Restnet} and Figure \ref{fig:footpath_conv}. For the FootpathVisionBD dataset, the training loss and accuracy curves indicate stable convergence for all three models; however, noticeable differences in learning efficiency and final performance are observed. The pretrained models, ResNet-50 and ConvNeXt-Tiny, exhibit a rapid decrease in training loss during the initial epochs, reflecting their ability to leverage pretrained visual representations learned from large-scale datasets. This results in earlier stabilization of accuracy compared to the custom CNN.

ConvNeXt-Tiny demonstrates smoother loss trajectories and more consistent accuracy improvement throughout training, suggesting improved robustness to the diverse urban scene variations present in the dataset, such as occlusions, varying lighting conditions, and heterogeneous object appearances. ResNet-50 also achieves competitive performance, although minor fluctuations in accuracy are observed, which may be attributed to its deeper architecture and higher sensitivity to dataset-specific noise. In contrast, the custom CNN converges more gradually and achieves slightly lower accuracy, indicating limitations in learning discriminative features solely from the available training samples. These results suggest that transfer learning provides a significant advantage for urban scene classification tasks, where complex spatial and contextual features are critical for accurate footpath encroachment detection.
    \vspace{0.3cm} 
 

\subsubsection{RoadDamageBD}
\begin{figure}[h]
    \centering
    \begin{subfigure}[b]{0.48\textwidth} 
        \centering
        \includegraphics[width=\textwidth]{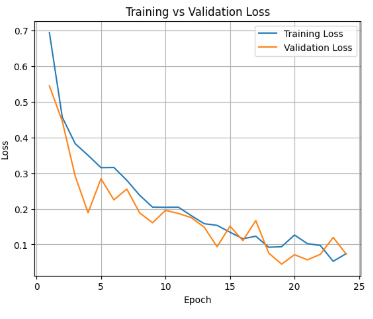}
        \caption{Custom CNN Loss} 
    \end{subfigure}
    \hfill
    \begin{subfigure}[b]{0.48\textwidth}
        \centering
        \includegraphics[width=\textwidth]{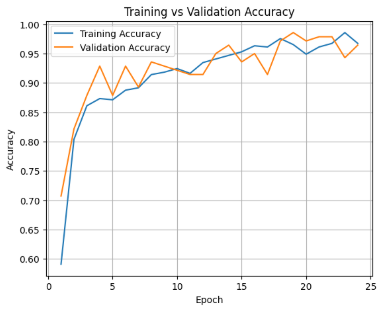}
        \caption{Custom CNN Accuracy}
    \end{subfigure}
    \caption{Training loss and Acuuracy curves of Custom CNN model across epochs on RoadDamageBD Dataset.}
    \label{fig:road}
\end{figure}

\begin{figure}[h]
    \centering
    \begin{subfigure}[b]{0.48\textwidth} 
        \centering
        \includegraphics[width=\textwidth, height=4.5cm]{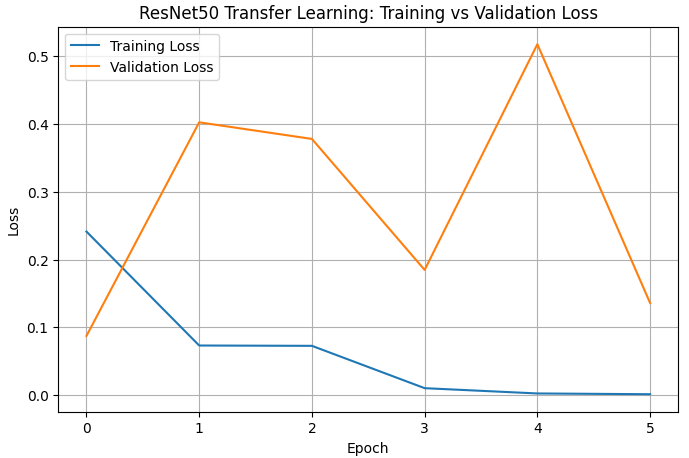}
        \caption{RestNet-50 Loss} 
    \end{subfigure}
    \hfill
    \begin{subfigure}[b]{0.48\textwidth}
        \centering
        \includegraphics[width=\textwidth, height=4.5cm]{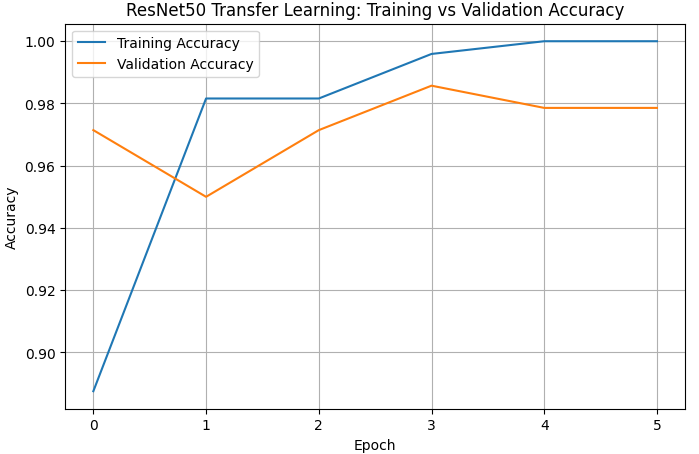}
        \caption{RestNet-50 Accuracy}
    \end{subfigure}
    \caption{Training loss and Acuuracy curves of RestNet-50 across epochs on RoadDamageBD Dataset.}
    \label{fig:road_Restnet}
\end{figure}

\begin{figure}[h]
    \centering
    \begin{subfigure}[b]{0.48\textwidth} 
        \centering
        \includegraphics[width=\textwidth, height=4.5cm]{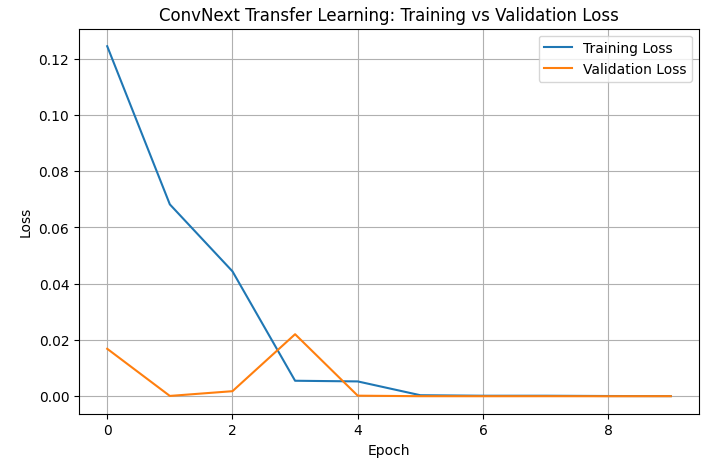}
        \caption{ConvNext-Tiny Loss} 
    \end{subfigure}
    \hfill
    \begin{subfigure}[b]{0.48\textwidth}
        \centering
        \includegraphics[width=\textwidth, height=4.5cm]{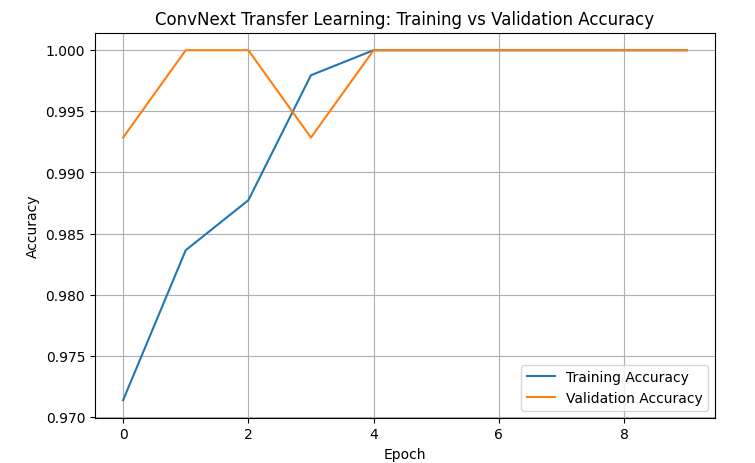}
        \caption{ConvNext-Tiny Accuracy}
    \end{subfigure}
    \caption{Training loss and Acuuracy curves of ConvNext-Tiny across epochs on RoadDamageBD Dataset.}
    \label{fig:road_conv}
\end{figure}

The dataset was already designed for binary classification, and after training for 20 epochs, 85.03\% accuracy was reached. The loss curve and accuracy curves are given in Figures ~\ref{fig:road},~\ref{fig:road_Restnet} and ~\ref{fig:road_conv}. For the RoadDamageBD dataset, the loss and accuracy curves indicate strong performance across all models. The pretrained models achieve rapid loss minimization and stable accuracy, highlighting their effectiveness in binary classification tasks with limited data. The custom CNN performs competitively but requires additional epochs to converge, suggesting that pretrained features are beneficial for capturing structural road damage patterns.

    \vspace{0.3cm} 
 

\subsubsection{MangoImageBD}
\begin{figure}[h]
    \centering
    \begin{subfigure}[b]{0.48\textwidth} 
        \centering
        \includegraphics[width=\textwidth, height=4.5cm]{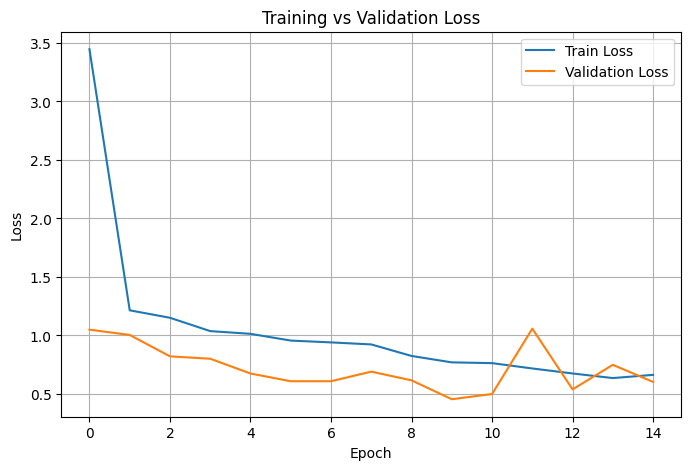}
        \caption{Custom CNN Loss} 
    \end{subfigure}
    \hfill
    \begin{subfigure}[b]{0.48\textwidth}
        \centering
        \includegraphics[width=\textwidth, height=4.5cm]{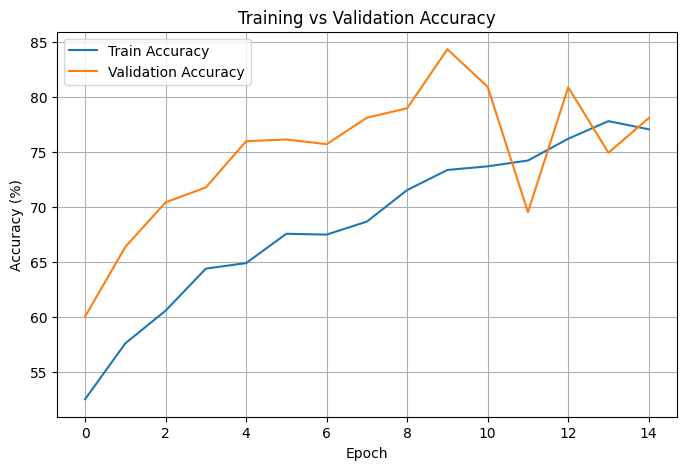}
        \caption{Custom CNN Accuracy}
    \end{subfigure}
    \caption{Training loss and Acuuracy curves of Custom CNN model across epochs on MangoImageBD Dataset.}
    \label{fig:mango}
\end{figure}

\begin{figure}[h]
    \centering
    \begin{subfigure}[b]{0.48\textwidth} 
        \centering
        \includegraphics[width=\textwidth, height=4.5cm]{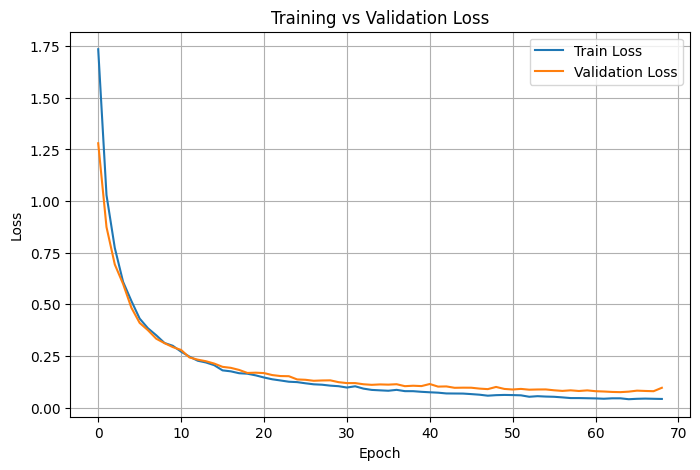}
        \caption{RestNet-50 Loss} 
    \end{subfigure}
    \hfill
    \begin{subfigure}[b]{0.48\textwidth}
        \centering
        \includegraphics[width=\textwidth, height=4.5cm]{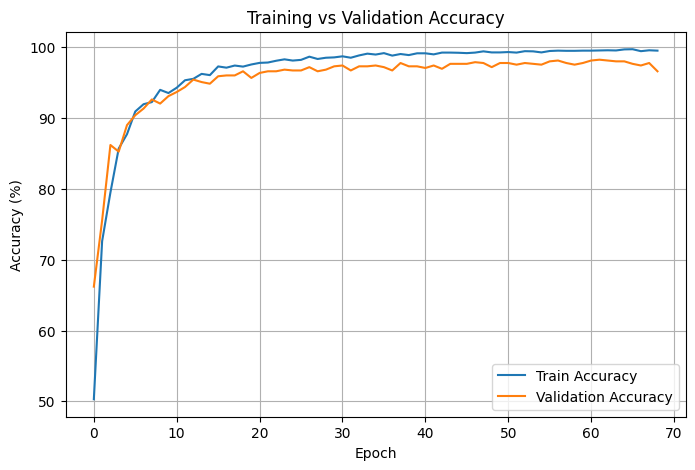}
        \caption{RestNet-50 Accuracy}
    \end{subfigure}
    \caption{Training loss and Acuuracy curves of RestNet-50 across epochs on MangoImageBD Dataset.}
    \label{fig:mango_Restnet}
\end{figure}

\begin{figure}[h]
    \centering
    \begin{subfigure}[b]{0.48\textwidth} 
        \centering
        \includegraphics[width=\textwidth, height=4.5cm]{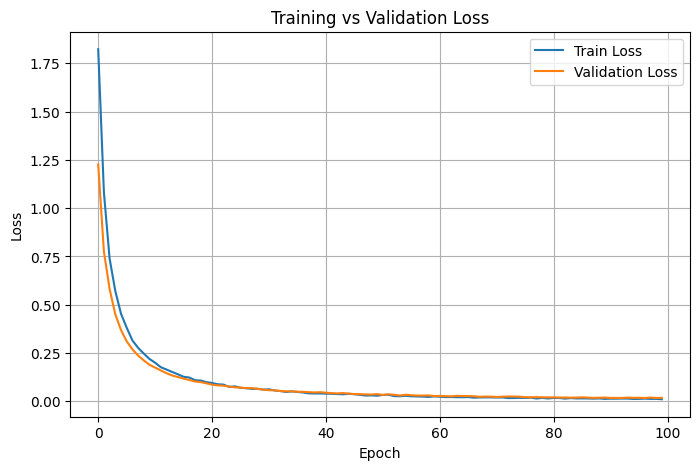}
        \caption{ConvNext-Tiny Loss} 
    \end{subfigure}
    \hfill
    \begin{subfigure}[b]{0.48\textwidth}
        \centering
        \includegraphics[width=\textwidth, height=4.5cm]{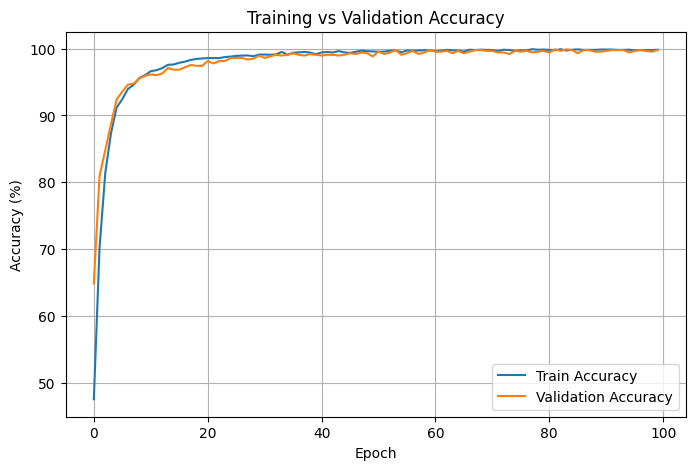}
        \caption{ConvNext-Tiny Accuracy}
    \end{subfigure}
    \caption{Training loss and Acuuracy curves of ConvNext-Tiny across epochs on MangoImageBD Dataset.}
    \label{fig:mango_conv}
\end{figure}

The dataset was already designed for multiclass classification, and after training for 20
epochs, 85.03\% accuracy was reached. The loss curve and accuracy curves are given
in Figures ~\ref{fig:mango}, \ref{fig:mango_Restnet} and ~\ref{fig:mango_conv}. For the MangoImageBD multiclass dataset, the performance gap between the pretrained models and the custom CNN becomes more pronounced. Both ResNet-50 and ConvNeXt-Tiny demonstrate faster convergence and more stable accuracy trends, indicating effective learning of fine-grained visual cues such as color, shape, and surface texture differences among mango varieties. The custom CNN exhibits slower convergence and minor accuracy fluctuations, suggesting difficulty in distinguishing closely related classes without deeper or pretrained feature extractors.
    \vspace{0.3cm} 
 

\subsubsection{PaddyVarietyBD}
\begin{figure}[h]
    \centering
    \begin{subfigure}[b]{0.48\textwidth} 
        \centering
        \includegraphics[width=\textwidth]{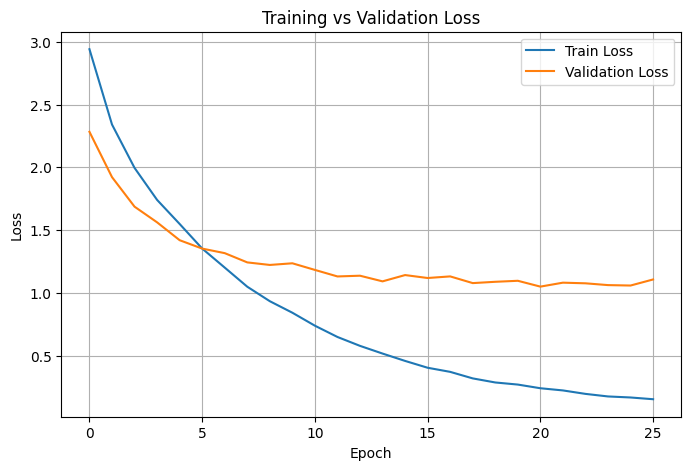}
        \caption{Custom CNN Loss} 
    \end{subfigure}
    \hfill
    \begin{subfigure}[b]{0.48\textwidth}
        \centering
        \includegraphics[width=\textwidth]{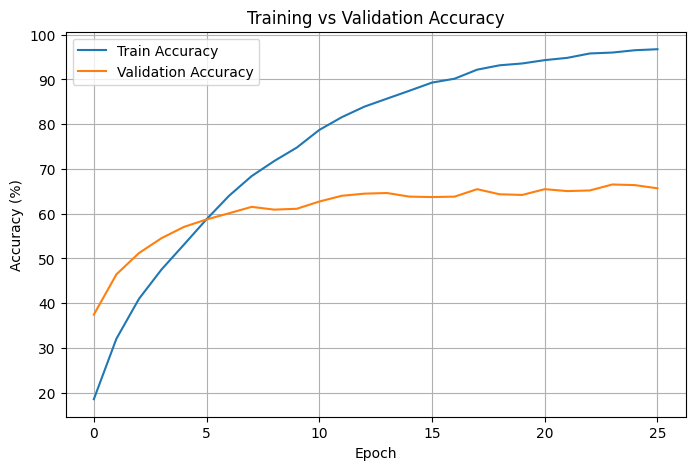}
        \caption{Custom CNN Accuracy}
    \end{subfigure}
    \caption{Training loss and Acuuracy curves of Custom CNN model across epochs on PaddyVarietyBD Dataset.}
    \label{fig:paddy}
\end{figure}

\begin{figure}[h]
    \centering
    \begin{subfigure}[b]{0.48\textwidth} 
        \centering
        \includegraphics[width=\textwidth]{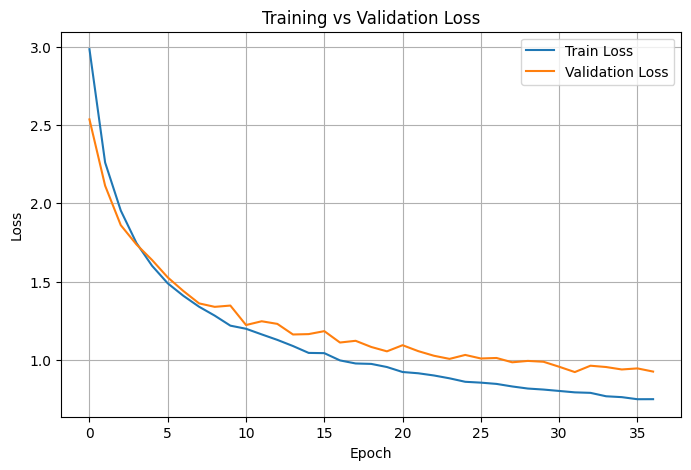}
        \caption{RestNet-50 Loss} 
    \end{subfigure}
    \hfill
    \begin{subfigure}[b]{0.48\textwidth}
        \centering
        \includegraphics[width=\textwidth]{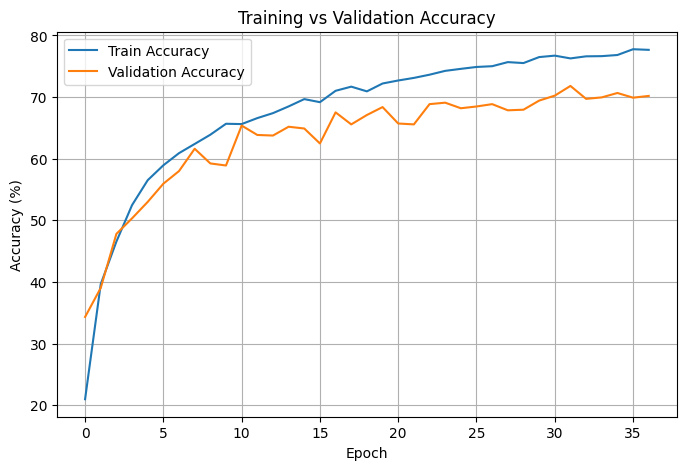}
        \caption{RestNet-50 Accuracy}
    \end{subfigure}
    \caption{Training loss and Acuuracy curves of RestNet-50 across epochs on PaddyVarietyBD Dataset.}
    \label{fig:paddy_Restnet}
\end{figure}

\begin{figure}[h]
    \centering
    \begin{subfigure}[b]{0.48\textwidth} 
        \centering
        \includegraphics[width=\textwidth]{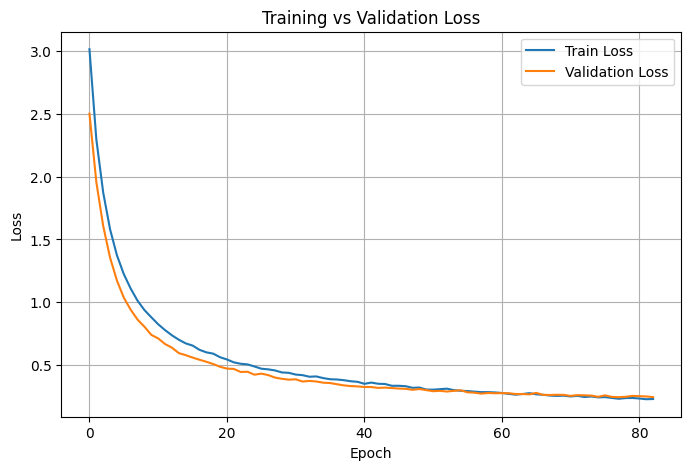}
        \caption{ConvNext-Tiny Loss} 
    \end{subfigure}
    \hfill
    \begin{subfigure}[b]{0.48\textwidth}
        \centering
        \includegraphics[width=\textwidth]{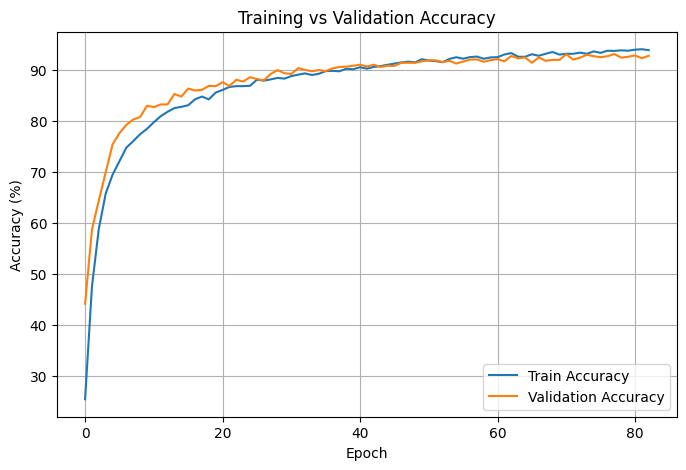}
        \caption{ConvNext-Tiny Accuracy}
    \end{subfigure}
    \caption{Training loss and Acuuracy curves of ConvNext-Tiny across epochs on PaddyVarietyBD Dataset.}
    \label{fig:paddy_conv}
\end{figure}

The dataset was already designed for multiclass classification, and after training for 20
epochs, 85.03\% accuracy was reached. The loss curve and accuracy curves are given
in Figures ~\ref{fig:paddy}, \ref{fig:paddy_Restnet}  and ~\ref{fig:paddy_conv}. The PaddyVarietyBD dataset presents a fine-grained multiclass classification challenge, where subtle texture and morphological differences are critical. ConvNeXt-Tiny achieves the most stable training behavior, with smooth loss reduction and consistently high accuracy, demonstrating its suitability for fine-grained visual recognition tasks. ResNet-50 also performs well but shows slight fluctuations during training, potentially due to its higher model complexity. The custom CNN converges more slowly and achieves comparatively lower accuracy, indicating limitations in capturing subtle inter-class variations when trained from scratch. This indicates the difficulty of the dataset compared to the other datasets and the importance of generalized knowledge from ImageNet dataset for better performance.
    \vspace{0.3cm} 
 

Across all datasets, the results consistently indicate that pretrained models outperform the custom CNN in terms of convergence speed, training stability, and final accuracy. In particular, ConvNeXt-Tiny provides a favorable balance between performance and efficiency, while ResNet-50 excels in feature richness at a higher computational cost. These findings emphasize the practical benefits of transfer learning, especially for small to medium-sized datasets and fine-grained classification tasks.

\subsection{Result Discussions}
The experimental results as shown in Table \ref{table:results-full} demonstrate consistent performance differences among the Custom CNN, ResNet-50, and ConvNeXt-Tiny models across the evaluated datasets. These differences can be primarily attributed to architectural design choices, the use of transfer learning, and the complexity of the datasets.

The Custom CNN, trained from scratch, shows comparatively slower convergence and slightly lower performance across most datasets. This behavior is expected, as the model must learn both low-level and high-level visual features solely from the target data. While the architecture is relatively shallow and computationally efficient, its limited representational capacity constrains performance, particularly for datasets involving fine-grained or multiclass classification. Nevertheless, the Custom CNN achieves competitive accuracy on simpler binary classification tasks, indicating that it is sufficient for problems with limited visual complexity and constrained resources.

In contrast, ResNet-50 consistently achieves higher accuracy and faster convergence, owing to its deep residual architecture and pretrained ImageNet weights. Residual connections facilitate stable gradient flow, enabling effective learning even in deeper networks. The pretrained features allow ResNet-50 to generalize well across diverse datasets, particularly those with complex visual patterns. However, minor fluctuations observed in some training curves can be attributed to the model’s higher parameter count, which increases sensitivity to dataset size and class imbalance.

ConvNeXt-Tiny demonstrates the most stable training behavior across datasets, with smooth loss reduction and consistent accuracy improvements. Its modernized convolutional design, inspired by Vision Transformer principles, enables efficient feature extraction while maintaining strong inductive biases. The relatively smaller model size compared to ResNet-50 allows ConvNeXt-Tiny to balance performance and computational efficiency, making it particularly effective for both binary and multiclass tasks.

Overall, the results indicate that transfer learning plays a crucial role in improving convergence speed, training stability, and generalization performance, especially for datasets with limited samples or fine-grained class distinctions. While the Custom CNN offers simplicity and reduced computational cost, pretrained architectures particularly ConvNeXt-Tiny provide a superior trade-off between accuracy and efficiency. These findings highlight the importance of selecting model architectures based on dataset complexity and available computational resources.

\begin{table}[ht]
    \centering
    \caption{Performance comparison of models across five datasets. Time is in minutes and parameter count includes only trainable parameters. Best scores in each dataset are \textbf{bolded}.}
    \begin{tabular}{|p{1.5cm}|p{2.5cm}|c|c|c|c|c|c|}
    \hline
    Dataset & Model & Accuracy & Precision & Recall & F1-score & Time & Parameters \\ \hline

    \multirow{3}{*}{\parbox{1.5cm}{Auto-Rickshaw Image BD}} & CustomCNN & 0.9204 & 0.9276 & 0.9204 & 0.912 & 10.02 & 134,608,642  \\
    & ResNet-50 & 0.791 & 0.754 & 0.791 & 0.7195 & 9.83 & 4,098  \\
    & ConvNeXT\_Tiny & \textbf{0.96} & \textbf{0.95} & 0.86 & 0.9 & 284.2 & 14,290,946 \\ \hline

    \multirow{3}{*}{\parbox{1.5cm}{Footpath Vision BD}} & CustomCNN & \textbf{0.9358} & \textbf{0.9369} & 0.9358 & 0.936 & 20.83 & 134,608,642  \\
    & ResNet-50 & 0.92 & 0.91 & \textbf{0.94} & \textbf{0.93} & 83 & 16,014,850  \\
    & ConvNeXT\_Tiny & 0.92 & 0.92 & 0.92 & 0.92 & 101.53 & 14,684,162 \\ \hline

    \multirow{3}{*}{\parbox{1.5cm}{Road Damage BD}} & CustomCNN & 0.9155 & 0.9292 & 0.9155 & 0.9158 & 42 & 13,250,241  \\
    & ResNet-50 & 0.9859 & 0.9864 & 0.9859 & 0.9859 & 20.07 & 17,221,121  \\
    & ConvNeXT\_Tiny & \textbf{1} & \textbf{1} & \textbf{1} & \textbf{1} & 231 & 20,367,361 \\ \hline

    \multirow{3}{*}{\parbox{1.5cm}{Mango Image BD}} & CustomCNN & 0.8481 & 0.8407 & 0.8481 & 0.8329 & 20.62 & 67,506,447  \\
    & ResNet-50 & 0.9755 & 0.9765 & 0.9755 & 0.975 & 60.51 & 30,735  \\
    & ConvNeXT\_Tiny & \textbf{0.9883} & \textbf{0.9887} & \textbf{0.9883} & \textbf{0.9883} & 126.34 & 13,071 \\ \hline

    \multirow{3}{*}{\parbox{1.5cm}{Paddy Variety BD}} & CustomCNN & 0.8971 & 0.8996 & 0.8971 & 0.8972 & 49.055 & 134,625,571 \\
    & ResNet-50 & 0.7257 & 0.7497 & 0.7257 & 0.7215 & 77.11 & 71,715 \\
    & ConvNeXT\_Tiny & \textbf{0.9195} & \textbf{0.9228} & \textbf{0.9195} & \textbf{0.9196} & 239.39 & 28,451 \\ \hline

    \end{tabular}
    \label{table:results-full}
\end{table}

Although the custom CNN contains significantly more trainable parameters, the pretrained ConvNeXt-Tiny achieves superior accuracy with fewer parameters, highlighting the efficiency of transfer learning.

\section{Conclusion}\label{sec5}
This study investigated the performance of deep learning approaches for both the binary and multiclass classification tasks. A progressive modeling strategy was applied, beginning with a custome convolutional neural network, followed by the use of two pre-trained models - ResNet-50 and ConvNeXt-Tiny as feature extractor and finally transfer learning through fine-tuning of the pre-trained models.

The evaluation of Custom CNN, ResNet-50, and ConvNeXT across five image datasets shows that ConvNeXT consistently achieves the highest accuracy and F1-scores, particularly on RoadDamageBD, MangoImageBD, and PaddyVarietyBD, demonstrating strong generalization across diverse datasets. CustomCNN performs competitively on FootpathVisionBD with lower computational requirements, while ResNet-50 generally excels in recall but lags in overall accuracy and F1-score. Overall, ConvNeXT is the most effective model in terms of predictive performance, though at the cost of longer training times and moderate parameter count, whereas CustomCNN and ResNet-50 offer favorable trade-offs between parameter efficiency and performance, making them suitable for scenarios with limited computational resources or recall-focused objectives.

\bibliography{sn-bibliography}

\end{document}